\documentclass{article}

\usepackage{arxiv}

\usepackage[utf8]{inputenc} % allow utf-8 input
\usepackage[T1]{fontenc}    % use 8-bit T1 fonts
\usepackage{verbatim}       % multi-line comments
\usepackage{makecell}       % multi-line cells in tables
\usepackage{hyperref}       % hyperlinks
\usepackage{url}            % simple URL typesetting
\usepackage{booktabs}       % professional-quality tables
\usepackage{amsmath}        % math typesetting
\usepackage{amsfonts}       % blackboard math symbols
\usepackage{nicefrac}       % compact symbols for 1/2, etc.
\usepackage{microtype}      % microtypography
\usepackage{authblk}        % simplyfy multiple authors from same institution
\usepackage{lipsum}         % generate paragraphs of lorem ipsum text
\usepackage{float}          % force table placement
\usepackage[backend=biber, style=numeric, sorting=ynt]{biblatex} % references
\usepackage{graphicx}     
\graphicspath{ {./figures/} }

\usepackage{array}
\usepackage{tabularx}
\newcolumntype{s}{>{\hsize=.5\hsize}X}  % `Y` column 0.5 times the size of an `X` column

\title{Beyond human subjectivity and error: a novel AI grading system}

\author{Alexandra Gobrecht $^1$
    \thanks{corresponding author: \texttt{alexandra.gobrecht@iu.org}}
}
\author{Felix Tuma $^1$
    \thanks{corresponding author: \texttt{felix.tuma@iu.org}}
}
\author{Moritz Möller}
\author{Thomas Zöller}
\author{Mark Zakhvatkin}
\author{Alexandra Wuttig}
\author{Holger Sommerfeldt}
\author{Sven Schütt}

\affil{IU International University of Applied Sciences}

\begin{document}

\maketitle

\def\thefootnote{1}\footnotetext{These authors contributed equally to this work}

\begin{abstract}
The grading of open-ended questions is a high-effort, high-impact task in education. Automating this task promises a significant reduction in workload for education professionals, as well as more consistent grading outcomes for students, by circumventing human subjectivity and error. While recent breakthroughs in AI technology might facilitate such automation, this has not been demonstrated at scale. It this paper, we introduce a novel automatic short answer grading (ASAG) system. The system is based on a fine-tuned open-source transformer model which we trained on large set of exam data from university courses across a large range of disciplines. We evaluated the trained model’s performance against held-out test data in a first experiment and found high accuracy levels across a broad spectrum of unseen questions, even in unseen courses. We further compared the performance of our model with that of certified human domain experts in a second experiment: we first assembled another test dataset from real historical exams – the historic grades contained in that data were awarded to students in a regulated, legally binding examination process; we therefore considered them as ground truth for our experiment. We then asked certified human domain experts and our model to grade the historic student answers again without disclosing the historic grades. Finally, we compared the hence obtained grades with the historic grades (our ground truth). We found that for the courses examined, the model deviated less from the official historic grades than the human re-graders - the model's median absolute error was $44\%$ smaller than the human re-graders', implying that the model is more consistent than humans in grading. These results suggest that leveraging AI enhanced grading can reduce human subjectivity, improve consistency and thus ultimately increase fairness.
\end{abstract}


\section{Introduction}
\label{sec:intro}
Grading exams and student papers (and especially grading answers to open ended questions) is one of the most laborious yet critically important tasks in higher education. Typically, this task is carried out by highly skilled human graders, i.e., professors, teaching assistants and tutors. While those graders usually are highly qualified academically and have substantial expertise within their subject area, they are still susceptible to the errors and biases that characterise human cognition \cite{mohlerTexttotextSemanticSimilarity2009,mohlerLearningGradeShort2011, smithImpactFramingEffect2009a}. This poses a significant risk to students, as their entire career might be impacted by the grades they receive. Further, grading duties consume a substantial fraction of the time of professors and tutors, which could have otherwise been spent on more direct teaching interventions. 

These problems can be partially mitigated through using multiple choice formats, which can be graded automatically and are hence less susceptible to bias and less time-intensive to grade. However, the expressiveness of multiple-choice formats is limited—they are typically supplemented by open-ended questions to probe student knowledge and skill more thoroughly—at the cost of the drawbacks described above. How the advantages of multiple-choice questions and open-ended questions can be combined is an open question.

A promising potential solution to this question is the application of Artificial Intelligence (AI) technology, especially the recent wave of large language models \cite{openaiGPT4TechnicalReport2024}. Several previous studies along these lines yielded encouraging results (see Section \ref{sec:backgrnd}); however, so far experiments have mostly been focused on small domains, which (in conjunction with other limitations) impeded scalability. In this paper, we present a novel automatic short-answer grading system (ASAG) for exam questions that overcomes these challenges. 

We start with reviewing previous attempts to automate the grading of open-ended questions with AI in Section \ref{sec:backgrnd}. We move on to outlining the design of our system and approach in Section \ref{subsec:setup}. We then put our system to test in two experiments, using real-world exam data for bachelor’s and master’s degrees. In the first experiment we evaluate the system using held-out test data; the results of this are reported in Section \ref{subsec:expone}. Subsequently, in a second experiment, we juxtapose our system with human graders to assess its performance against a human benchmark in Section \ref{subsec:exptwo}. Finally, we discuss the implications of our findings in Section \ref{sec:disc}, where we lay out a high-level road map for autonomous grading in real-world applications.

The findings in this study are encouraging; in summary, at the time of writing and as far as the authors are aware, our system does not only utilize the largest training set ever assembled and described in the automatic grading domain but also holds a unique position as the first system that, when compared and benchmarked against human graders, outperforms them. This signifies a substantial leap forward in the pursuit of trustworthy, unbiased grading systems.

\section{Background}
\label{sec:backgrnd}
The problem of grading open-ended questions can be viewed as a regression problem: based on several independent variables $X$ (including, but not limited to, the question and the student’s answer), we try to predict a dependent variable $y$ (the grade). Often, the independent variable is text, while the dependent variable is numerical—for example the number of points that are awarded for an answer. This setup allows to apply common methods of evaluation, such as Pearson correlation coefficients, RMSE and MAE (for explanations of these metrics, see Section \ref{subsec:evalmetrics}). With that problem statement in mind, we now review several previous solution attempts in the following paragraphs.

In one of the first attempts to tackle automated grading of open-ended questions, \cite{mohlerTexttotextSemanticSimilarity2009} evaluated various measures of word and text similarity for automatic short answer grading, as well as corpus-based measures using different corpora such as Wikipedia and BNC (British National Corpus). They also proposed a technique incorporating paraphrases of student answers—formulations that should obtain the same grade although using different words—into grading, to improve the performance of their system. Their dataset consisted of introductory computer science assignment questions with answers from undergraduate students, with a maximum achievable grade of 5 points for each question. They observed that grading short answer tasks can be subjective, with only 56.8\% of grades being in exact agreement with human annotators. 17.0\% of grades differed by more than one point on the five-point scale, and 3.0\% differed by 4 points or more. Overall, their best system includes a corpus-based measure (LSA, Latent Semantic Analysis \cite{landauerSolutionPlatoProblem1997}) trained on a domain-specific corpus built on Wikipedia with feedback from student answers. This system shows a significant absolute improvement of 0.14 points on the Pearson scale (absolute Pearson's correlation coefficient of 0.5099 points) over the tf*idf baseline and 0.10 points over another LSA model trained only on the BNC corpus. 

In a follow-up study, \cite{mohlerLearningGradeShort2011} expanded their dataset and combined graph alignment features, semantic similarity measures, and machine learning techniques for aligning dependency graphs of both student and instructor answers. This alignment allowed the insertion of a structural component in the automated grading, which lead to improved grade learning. The authors report a Pearson's correlation coefficient of 0.518 and RMSE of 0.998 points for the best model version. For the inter annotator agreement (agreement between the two human graders), they report a Pearson's correlation of 0.586 points and an RMSE 0.659 points. This means that their best model is not able to perform as well as the two human graders.

Another approach was taken by \cite{schlippeDeepLearningTechniques}, who modified a BERT-based model using linear regression for English and German, with a German dataset from an online exam system with the maximum number of points ranging from 6 to 10 points. The authors report a Pearson correlation coefficient of 0.73 points, an RMSE of 0.72 points and an MAE of 0.42 points on the Short Answer Grading data set of\cite{mohlerLearningGradeShort2011}, thus improving upon the best previous model. On the German dataset, \cite{schlippeDeepLearningTechniques} report a Pearson correlation coefficient of 0.78 points, an RMSE of 1.62 points and an MAE of 1.19 points.

More recently, \cite{pintoLargeLanguageModels2023} explored the potential of ChatGPT as a novel approach for correcting answers to open-ended questions while also offering direct feedback. ChatGPT was used to correct open-ended questions that were answered by 42 professionals in the industry, with a focus on two specific topics (web application caching, and stress and performance testing). The consensus among the two experts was that the corrections suggested by ChatGPT were accurate. Out of the six feedback statements provided by ChatGPT, there was only one instance where the expert disagreed with ChatGPT. The researchers also noted that ChatGPT demonstrated the ability to identify semantic details in responses that were not captured by other evaluation metrics. As the authors did not use the typical quantitative metrics (Pearson coefficients, RMSE, MAE) to evaluate their approach, their results are not comparable to the other approaches or our own approach.

The results of these previous works are promising and important, as they prove the viability of the AI approach for grading. However, the above-mentioned approaches have a critical shortcoming: they are not designed for, nor have they been tested on, unseen questions, let alone in unseen domains. This constitutes a serious limitation for their application in practise—in real-world scenarios, the domains of application and the questions that are being graded are typically highly dynamic. For example, exam questions are often changed every time a course is being taught, to prevent cheating. Further, the experiments described above were often limited to small domains of knowledge—it is hence unclear whether the obtained results generalize to other domains.

The system we present in the following sections addresses these issues. It is based on a novel automatic short-answer grading model—ASAG for short—which most closely resembles the approach of \cite{schlippeDeepLearningTechniques}. However, compared to previous works, we use a substantially larger training set and updated designs—also encompassing unseen questions—as well as a wider variety of evaluation methods, including a human benchmark experiment. The details of our model setup and training, the results of the test-set based model evaluation (experiment 1) and the results of the human benchmark experiment (experiment 2) are described in the next section.

\section{Results}
\label{sec:results}
This section describes the setup and evaluation of our automated short answer grading (ASAG) model. It consists of three parts: in the first part \ref{subsec:setup} we describe the system architecture and training approach. In the second part \ref{subsec:expone}, we describe the evaluation of the model based on held-out test sets and report the results—we will also refer to this as experiment 1. In the third part \ref{subsec:exptwo}, we compare human and model grading in a human-benchmark experiment—this is referred to as experiment 2. 

\subsection{Model setup and training}
\label{subsec:setup}
In this subsection, we outline the system architecture and then describe our finetuning approach.

\paragraph{System Architecture}
As the core of our system, we first selected a large open-source transformer model. For the use case of grading open-ended questions, we then defined the inputs for the model to be a tuple consisting of the following 4 elements: 

\begin{enumerate}
    \item A question, $Q$
    \item A reference answer that would receive full marks, $A_{\text{ref}}$
    \item The maximum number of points achievable in the question, $x_{\text{ref}}$
    \item The student answer to be evaluated, $A$
\end{enumerate}

\begin{figure}
    \centering
    \includegraphics[width=.8\textwidth]{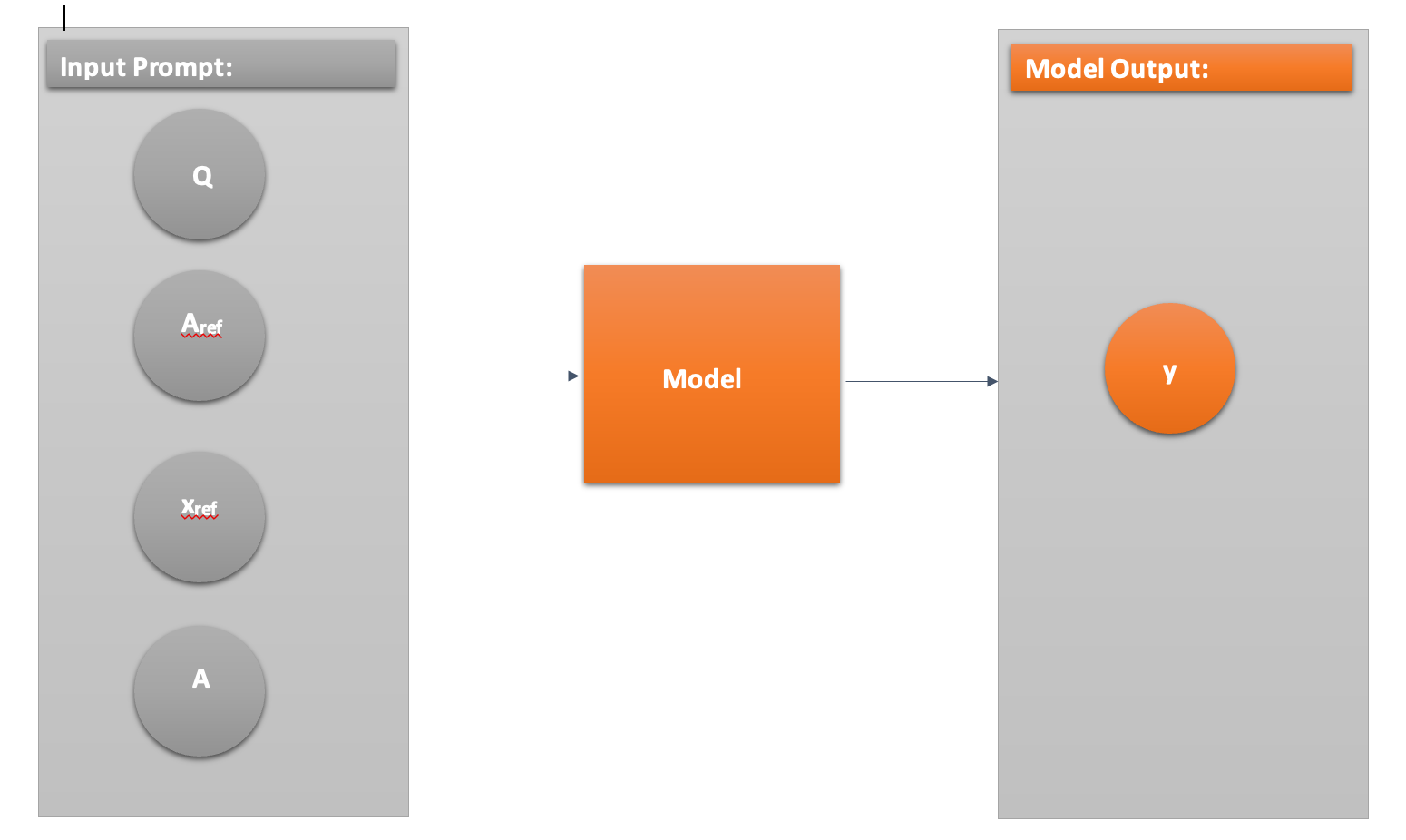}
    \caption[Model inputs and outputs] {\textbf{Model inputs and outputs.} The model receives a question, a reference answer, the student answer and the maximum number of points as inputs and provides a grade (the number of points for the student answer) as an output.}
    \label{fig:modelinout}
\end{figure}

The output $y$ was defined to be the grade—the number of points that the answer to be evaluated, $A$, would receive. These inputs and outputs of the system are visualized in Fig \ref{fig:modelinout}, an example input-output pair is given in Table \ref{tab:exinout}. Note that a sensible output $y$ of the model should fulfil $y <= x_{\text{ref}}$, but this is not enforced by the architecture. 

\begin{table}[htbp]
    \caption[Example Input Output Pair]{\textbf{Example Input Output Pair.} Here, the students answer received 4 out of 6 possible points.}
    \centering
    \begin{tabularx}{\textwidth}{sX}
        \toprule
        \textbf{Input} & \\
        \midrule
        \textbf{Question} $Q$ & Describe the effectiveness and efficiency approaches. \\
        \textbf{Reference answer} $A_\text{ref}$ & Efficiency describes the ratio of output and input quantities evaluated in monetary  units,  i.e. performance rendered and use of resources. Effectiveness assesses target achievement,  i.e. to what extent the planned targets were also achieved. \\
        \textbf{Maximum points} $x_{\text{ref}}$ & 6.0 points \\
        \textbf{Student Answer} $A$ & Effectiveness refers to whether a goal has been achieved with the chosen means, e.g.: a fire  should be extinguished - you use water, the fire goes out - the agent (water) was effective. Efficiency refers to whether the chosen funds are profitable. For example, I can also  extinguish the fire with expensive wine, but that wouldn't be efficient. \\
        \midrule
        \textbf{Output} & \\
        \midrule
        \textbf{Grade} $y$ & 4.0 points \\
        \bottomrule
    \end{tabularx}
    \label{tab:exinout}
\end{table}

\paragraph{Fine-tuning Approach}
Using the above setup, we created a dataset of input-output pairs to fine-tune the model. This dataset was assembled from graded university exam questions and responses across various disciplines taught at the IU International University of Applied Sciences. As required by our system design described above, each record in this proprietary dataset includes a question, the student's answer, a high-quality reference answer, the maximum achievable points for that question (ranging from 6 to 18 points), and the actual points awarded in the official examination process. The reference answers were created by domain experts (i.e. tutors and professors) and were therefore be considered as ground truth. In some cases, the reference answers were annotated to guide grading (for example, specifying “(2 points)” after a part of the answer), but most reference answers did not have this feature—see Table \ref{tab:exinout} for a typical example without annotation.

This dataset included a broad spectrum of disciplines, from humanities to STEM fields, to promote the model's adaptability and accuracy across different subjects. This diverse compilation ensures the model's exposure to a wide array of grading standards, question formats, and subject matter expertise. We partitioned our dataset into three primary splits: 
$S_{\text{train}}$, $S_{\text{develop}}$, and $S_{\text{test}}$ (the latter being further divided into  
$S_{\text{test, unseen questions}}$ and $S_{\text{test, unseen courses}}$). These splits are described in detail in Section \ref{subsec:datapart} below, where we also provide statistics on the number of courses and questions contained in each split.

Using the above-described base model and the training split $S_{\text{train}}$, we performed supervised fine-tuning to increase the model’s capabilities in the grading domain. During fine-tuning, we measured the learning progress with the $S_{\text{develop}}$ split (1.8\% of data - unseen questions). Fine-tuning was continued until the learning curve had plateaued. The resulting fine-tuned model was then evaluated in two experiments, as described in the next sections.

\subsection{Experiment 1: Evaluation on held-out test sets}
\label{subsec:expone}
In this subsection, we describe our first method of evaluating our model, which consisted in using the held-out test sets $S_{\text{test, unseen questions}}$ and $S_{\text{test, unseen courses}}$ to determine the model’s performance and generalization abilities. We also provide the final evaluation metrics for the development split $S_{\text{develop}}$; however, since this split was already used for monitoring in the training process, the resulting metrics should only be taken as a complement to the actual test set metrics.

First, we focus on raw grades (the direct output of the model, i.e. points between $0$ and $18$). Second, we investigate normalized grades (the raw grade divided by the maximal number of points, resulting in a number between $0$ and $1$, i.e. a percentage). The results from these two analyses are summarized in Table \ref{tab:sumexpone}. We show various regression metrics for several different splits of the dataset.

\begin{table}[htbp]
    \caption[Summary of the results of experiment 1]{\textbf{Summary of the results of experiment 1.} We show various regression metrics for several different splits of the dataset.}
    \centering
    \label{tab:sumexpone}
    \begin{tabular}{lccc}
        \toprule
        & $S_{\text{develop}}$ & $S_{\text{test, unseen questions}}$ & $S_{\text{test, unseen courses}}$ \\
        \midrule
        MAE (points) & 1.4009 & 1.3207 & 1.4374 \\
        RMSE (points) & 2.3574 & 2.2704 & 2.4132 \\
        Correlation & 0.7295 & 0.7847 & 0.6945 \\
        \midrule
        MAE (normalized) & 0.1719 & 0.1559 & 0.1856 \\
        RMSE (normalized) & 0.2820 & 0.2558 & 0.2904 \\
        Correlation (normalized) & 0.6394 & 0.6143 & 0.6400 \\
        \bottomrule
    \end{tabular}
    
\end{table}

Third, we introduce another perspective by grouping our results by the maximal number of achievable points, which constitute a proxy for question (and hence grading) complexity—this way, we can explore how performance and complexity are related. For each of those analyses, performance was evaluated on all three splits, based on Mean Absolute Error (MAE), Root Mean Square Error (RMSE), and Pearson’s correlation coefficients (see Section \ref{subsec:evalmetrics} for an explanation of these metrics).

\paragraph{Raw Grade Performance}
For the $S_{\text{develop}}$ split, we measured an MAE of 1.4009 points and an RMSE of 2.3574 points, with a correlation of 0.7295 between predicted and actual point values, suggesting a strong alignment with human grading. In $S_{\text{test, unseen questions}}$, the MAE slightly decreased to 1.3207 points, and the RMSE to 2.2704 points, complemented by a higher correlation of 0.7847. Finally, $S_{\text{test, unseen courses}}$ showed an MAE of 1.4374 and an RMSE of 2.4132, with a correlation of 0.6945, indicating a slight decrease in grading accuracy, likely due to unfamiliar course content. Overall, the raw grade measurements do not vary strongly between the three splits.

\paragraph{Normalized Grade Performance}
When running the same evaluation for normalized grades, we found an MAE of 17.19 percent-points and an RMSE of 28.20 percent-points for $S_{\text{develop}}$. The correlation measured was 0.6394. The performance on $S_{\text{test, unseen questions}}$ was slightly better than that on $S_{\text{develop}}$, with a lower MAE of 15.59 percent-points and an equally lower RMSE of 25.58 percent-points, although the correlation decreased to 0.6143. The $S_{\text{test, unseen courses}}$ split set recorded an MAE of 18.56 percent-points and an RMSE of 29.04 percent-points, with a correlation of 0.6400, underscoring the difficulty in adjusting to new course material. Again, the values of our measures are mainly consistent across the different splits. 

\paragraph{Performance Split by Maximum Number of Points}
As a last step of analysis in experiment 1, we repeated the above analyses (raw and normalized grades) for each type of question separately—with types of questions defined by the maximum number of points that can be achieved. The results of this are given in tables \ref{tab:testunseenq} and \ref{tab:testunseenc}.

\begin{table}[htbp]
    \centering
    \caption{\textbf{Results on $S_{\text{test, unseen questions}}$ grouped by maximum number of points.}}
    \label{tab:testunseenq}
    \begin{tabular}{cccc}
        \toprule
        Test Unseen Questions & Percentage of Data & MAE (normalized) & Std \\
        Max number of points & & & \\
        \midrule
        6 & 41.2\% & 0.1490 & 0.2113 \\
        8 & 26.4\% & 0.1590 & 0.2021 \\
        10 & 25.4\% & 0.1751 & 0.2040 \\
        18 & 7\% & 0.2514 & 0.2402 \\
        \bottomrule
    \end{tabular}
\end{table}

\begin{table}[htbp]
    \centering
    \caption{\textbf{Results on $S_{\text{test, unseen courses}}$ grouped by maximum number of points.}}
    \label{tab:testunseenc}
    \begin{tabular}{cccc}
        \toprule
        Test Unseen Courses & Percentage of Data & MAE (normalized) & Std \\
        Max number of points & & & \\
        \midrule
        6 & 59.1\% & 0.1797 & 0.2320 \\
        8 & 18.1\% & 0.1874 & 0.2125 \\
        10 & 19.4\% & 0.1837 & 0.1926 \\
        18 & 3.4\% & 0.2890 & 0.2624 \\
        \bottomrule
    \end{tabular}
\end{table}

We found a clear trend: the mean error increased with higher maximum grades across both test sets. Notably, the mean error for questions with a maximum grade of 18.0 was substantially higher than for questions with lower maximum grades, suggesting that the model's grading accuracy diminishes with increasing question complexity.

This effect might be partially due to the under-representation of questions with a maximum number of points of $18$ in the training set—as the model's exposure to high-complexity questions was limited, its ability to accurately grade such items was impacted. Another possible explanation for the effect consists in observing that graders had a larger variety of options for questions with high maximum numbers of points. For example, consider a hypothetical question with a maximum number of points of $1$: Assuming that typically only integer points are given, such questions amount to binary classification problems, which are much easier than regression problems.

Overall, we find that our model generalises well to unseen questions and even completely unseen courses: across all modalities, we found mean absolute errors below 1.5 points per question (raw grade), which translates into absolute deviations below 20\% on average (normalised grade). While these metrics suggest that our approach is viable and generalizable, it is difficult to interpret them with regards to practical application—in other words, it is not yet clear whether the model is ‘good enough’ to serve in real world scenarios. Further benchmarks are needed for this; one benchmark of particular relevance is the human benchmark—how much would humans deviate from the original grade if they graded the questions in the test set again? To answer this question, we ran a second experiment, which we describe in the next section.

\subsection{Experiment 2: Comparing human against model performance in a grading task}
\label{subsec:exptwo}
After having outlined our modelling approach in Section \ref{subsec:setup} and reporting the evaluation of our model on unseen questions and courses in Section \ref{subsec:expone}, we now move on to compare our model’s performance with human performance. Roughly, we ask: is our model better or worse than human professionals in regrading open-ended questions? 

\paragraph{Comparing the Model to Human Re-Graders}
To test whether the model is better or worse than humans in grading, we took 100 question-answer pairs from each of 16 different courses (1600 pairs in total; sampled from the historical exam data test set described above in Section \ref{subsec:setup} —see Appendix \ref{sec:app:regrading} for the titles of the selected courses). As described above, those pairs had been part of official exams completed by students and graded by human professors and tutors, generating an official, legally binding exam grade, obtained in a strongly regulated and controlled process. Acknowledging the rigours of this process (and the fact that the resulting grades are generally accepted by employers and institutions), we consider them as the ground truth for this experiment, henceforth referred to as ‘official’ exam grades. 

\begin{figure} 
    \centering
    \includegraphics[width=.8\textwidth]{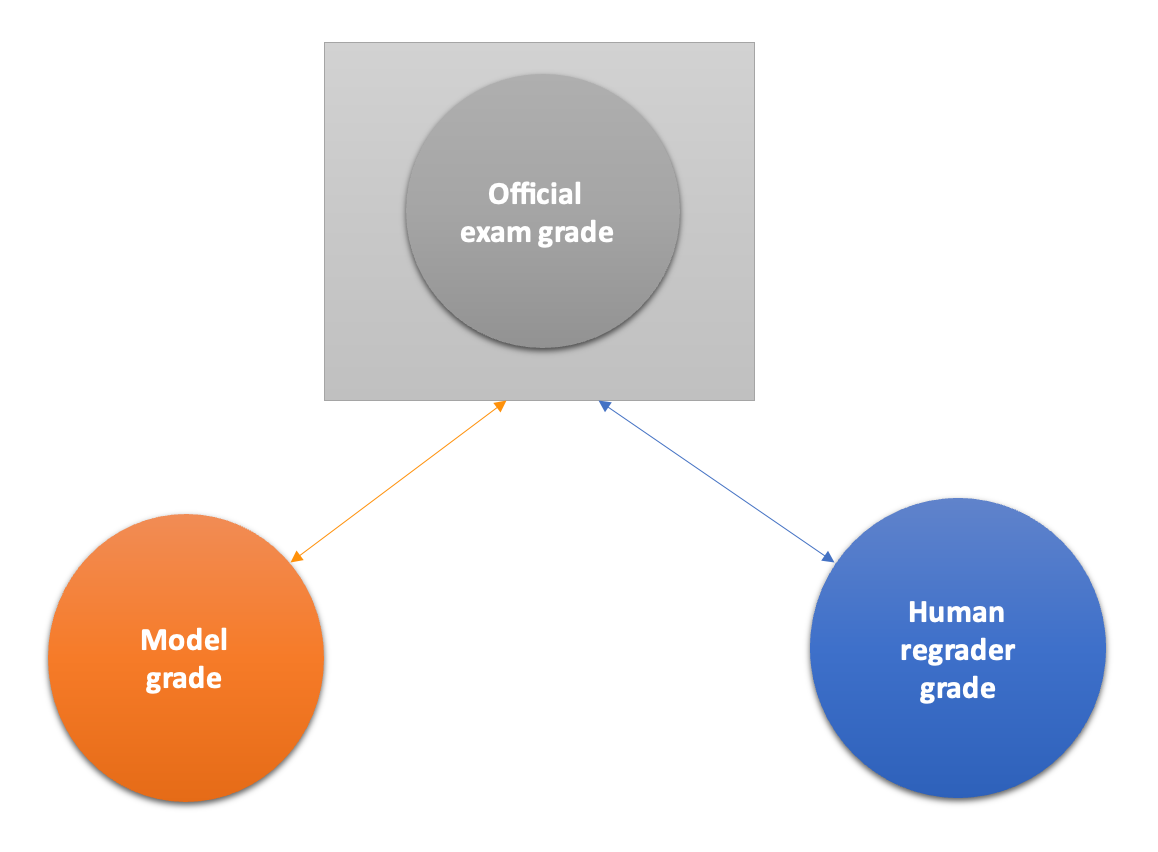}
    \caption[Visualisation of the relevant scores] {\textbf{Visualisation of the relevant scores.} We compared the deviation of model grades (left) and human regrader grades (right) from official exam grades (middle).}
    \label{fig:relscores}
\end{figure}

We then recruited four human annotators with domain expertise to regrade those question-answer pairs without seeing the official grade. Those annotators were tutors from the same university that generated the exam dataset and were qualified to grade exams in various courses – i.e., domain experts. Each of those four tutors focused on four courses that they were qualified for but had not yet been grading exams in, generating a human re-grader grade. Finally, we also had our model grade these 1600 rows, resulting in a model grade for each question-answer pair. All in all, we thus assembled three grades for each question-answer pair (see Fig \ref{fig:relscores} for a visualisation) — the official grade, the model grade and the human re-grader grade. We then compared the deviations between model and official grade on the one hand, and human re-grader and official grade on the other (conceptually, we treated the human re-grader grades similar to grades generated by a competitor model in a model comparison).

We found that overall, the model’s deviation from the official exam grades is substantially smaller than the human re-grader’s deviation from the official grades—in other words, human re-graders deviated more strongly than our model from the official grades. The human re-graders achieved an RMSE of 0.412 (4.566 points), a Pearson's correlation of 0.485 (0.583 points) and a mean deviation of 0.289 percentage points. The model achieved an RMSE of 0.284 (3.061 points), a Pearson's correlation of 0.590 (0.761 points) and a mean deviation of 0.183 percentage points.

\begin{figure} 
    \centering
    \includegraphics[width=.95\textwidth]{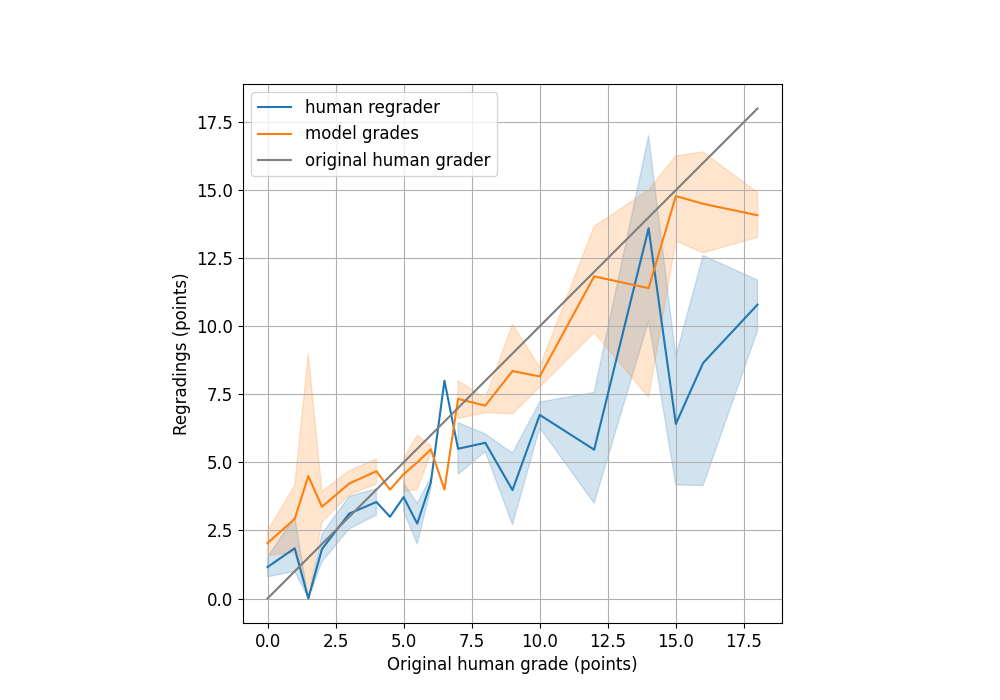}
    \caption[Results of the human benchmark experiment] {\textbf{Results of the human benchmark experiment.} We show all three grades (green: official grade – considered to be the ground truth, blue: human regrader grade, orange: model grade) on the y-axis, mapped against the official grade on the x-axis. The model grades are closer to the official grades (the ground truth) than the human re-grader grades.}
    \label{fig:humbench}
\end{figure}

Focusing on mean absolute deviation, we found that the model deviates 10.6 percentage points less compared to the human re-grader -- substituting the human by the model would reduce the deviation by 36.7\%. See Fig \ref{fig:humbench} and Table \ref{tab:sumdev} and for our main results and Table \ref{tab:app:gradvsregradone} in Appendix \ref{sec:app:detres} for detailed statistics of the experiment.

When looking at median absolute deviation (potentially more robust compared to the mean, as the distribution is strongly skewed), we find a value of $0.11$ percentage points for the model, and a value of $0.2$ percentage points for the human re-graders (see Table \ref{tab:sumdev}, $50\%$ column). This means that the model deviates $0.09$ percentage points less from the official grade than the human regraders - a reduction of the median absolute deviation by $44\%$. This key result is visualised in Fig \ref{fig:app:graderror}.

On a course level, there is one case out of 16 in which the human re-grader deviated less than the model: Schuldrecht I, Einführung (Law of obligation I, Introduction), with an RMSE of 0.294 (2.416 points) for the human versus 0.326 (2.652 points) for the model, Pearson's correlation of 0.619 versus 0.538. In all other courses (15 of 16), the model deviated less than the human re-graders.

\begin{table}[htbp]
    \caption[Summary of deviations]{\textbf{Summary of deviations.} Deviation between the two human graders d(hh) compared to deviation between model and official human exam grader d(hm), both in percentiles. For example, the ‘50\%’ column can be read as ‘In 50\% of cases, the model deviated less than 11\% from the official grade, while the human regraders deviated less than 20\%’.}
    \centering
    \label{tab:sumdev}
    \begin{tabular}{cccccccccc}
        \toprule
        & & & & & & & & & \makecell{\# of rows graded: 1600} \\
        \midrule
        & Mean & Std & Min & 25\% & 50\% & 75\% & 90\% & 95\% & Max \\
        \midrule
        d(hh) & 0.2886 & 0.2937 & 0.0000 & 0.2000 & 0.2000 & 0.5000 & 0.7500 & 1.0000 & 1.0000 \\
        d(hm) & 0.1830 & 0.2178 & 0.0000 & 0.0000 & 0.1111 & 0.3333 & 0.5000 & 0.5556 & 1.0000 \\
        \bottomrule
    \end{tabular}
\end{table}

In summary, these results suggest that the model is typically more in line with the official grade than human regraders: the model’s deviation is substantially smaller than the human regrader’s deviation from the official grade in 15 of 16 courses as well as across all courses combined. 

\begin{figure} 
    \centering
    \includegraphics[width=.95\textwidth]{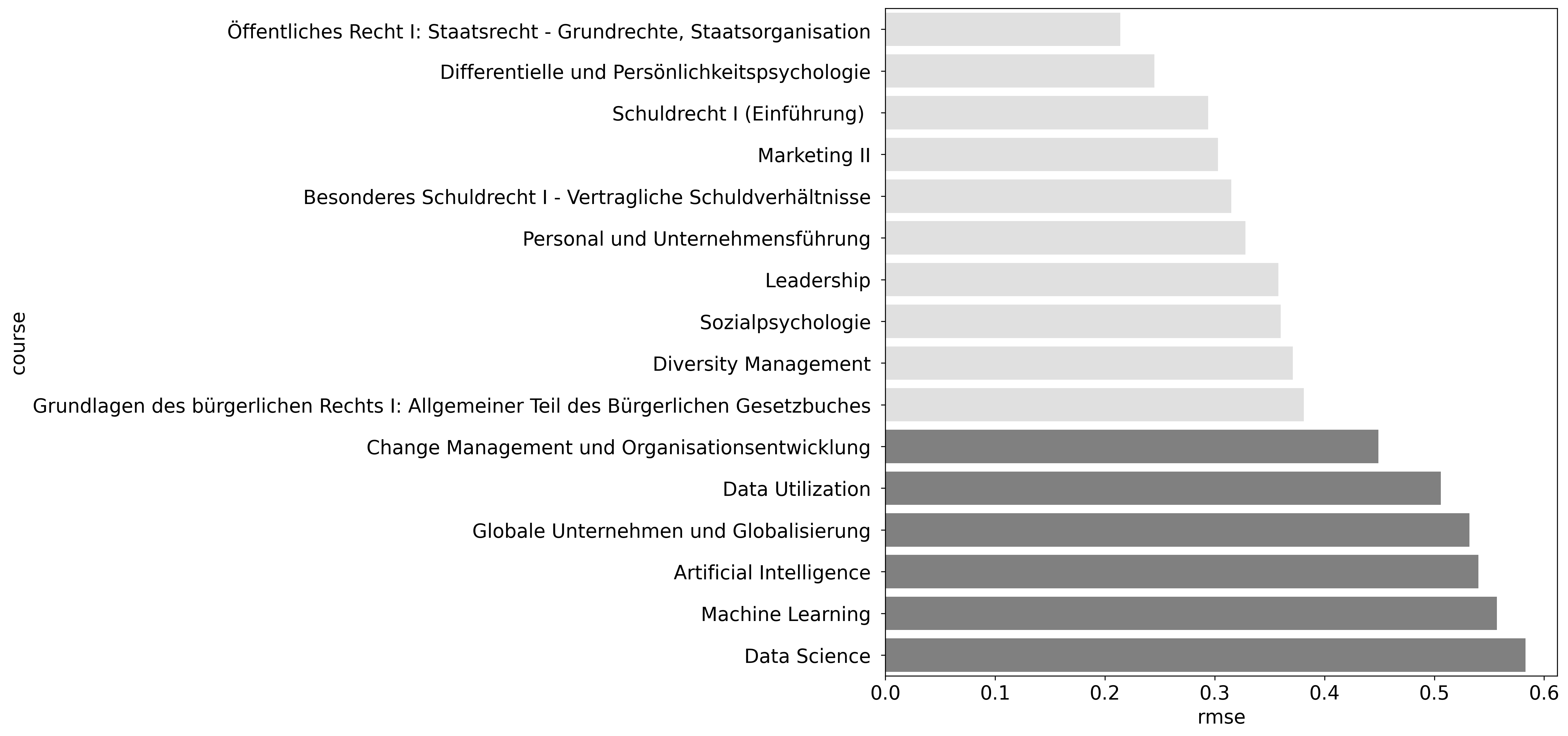}
    \caption[Re-grader deviations per course] {\textbf{Re-grader deviations per course.} We show RMSE between human re-graders and official grades per course. The ‘extreme’ re-graders that we excluded from the analysis in this section are coloured in dark grey. }
    \label{fig:devpercourse}
\end{figure}

\paragraph{Dealing with strongly deviating re-graders}
Some human re-graders had an extreme deviation from the original grader in six courses (up to RMSE 0.583 and Pearson's correlation coefficient of 0.308): \textit{Artificial Intelligence}, \textit{Data Utilization}, \textit{Data Science}, \textit{Change Management und Organisationsentwicklung (Change management and developing organisations)}, \textit{Globale Unternehmen und Globalisierung (Global enterprises and globalisation)}, and \textit{Machine Learning} -- see Fig \ref{fig:devpercourse} for a comparison of the re-graders. In a second analysis, we excluded these courses and ran a tight analysis on the remaining 1000 question-answer pairs to ensure that our results were not due to a minority of persons mistakes. Still, with the 1000 rows left, we found that across all these courses, the deviation of the human grader and the model is smaller than the deviation between the two human graders. On a single course level, the model deviation is smaller than human re-grader’s deviation in all courses except for one (\textit{Schuldrecht I, Einführung (Law of obligation I, Introduction)}).

After filtering out extreme re-graders, the deviation between the two human graders has an RMSE of 0.321 (2.926 points) and a Pearson's correlation coefficient 0.589 (0.643 points), the one between the human and our model has an RMSE of 0.279 (2.346 points) and a correlation of 0.600 (0.741 points). The mean deviation between the two human graders is now 0.2206 percentage points (compared with 0.289 before filtering out extreme re-graders), the one between the first human grader and the model is now 0.1773 (compared with 0.183 before). This means that the model deviates 0.0433 percentage points less from the official grade — substituting the non-extreme human re-graders by the model would reduce the deviation by 19.5\%.

\begin{table}[htbp]
    \caption[Excluding human outliers from the benchmark experiment]{\textbf{Excluding human outliers from the benchmark experiment.} Deviation between the two human graders d(hh) compared to deviation between model and official human exam grader d(hm), both in percentiles. The data shown here is similar to Table \ref{tab:sumdev} above, but with extreme human re-graders excluded.}
    \centering
    \label{tab:humout}
    \begin{tabular}{cccccccccc}
        \toprule
        & & & & & & & & & \makecell{\# of rows graded: 1000} \\
        \midrule
        & Mean & Std & Min & 25\% & 50\% & 75\% & 90\% & 95\% & Max \\
        \midrule
        d(hh) & 0.2206 & 0.2334 & 0.0000 & 0.0000 & 0.1667 & 0.3333 & 0.5000 & 0.6842 & 1.0000 \\
        d(hm) & 0.1773 & 0.2150 & 0.0000 & 0.0000 & 0.1000 & 0.3333 & 0.5000 & 0.5008 & 1.0000 \\
        \bottomrule
    \end{tabular}
\end{table}

In summary, even when excluding these outliers, the overall deviation of the human grader and the model is smaller than the deviation between the two human graders. Further, in all courses except for Schuldrecht I, Einführung (Law of obligation I, Introduction), the deviation of the human grader and the model is still significantly smaller than the deviation between the two human graders.

\section{Discussion}
\label{sec:disc}
Above, we started by introducing the problem of grading open-ended questions and continued by reviewing previous work on the topic in section 2. We then described the ASAG model in section \ref{subsec:setup} and evaluated its performance in section \ref{subsec:expone} (experiment 1). We further presented a comparison with human performance in section \ref{subsec:exptwo} (experiment 2), where we showed that ASAG is closer to the official reference grade than human re-graders on average, even when excluding strongly deviating re-graders. Here, we move on to discuss the ramifications of those results, especially regarding potential future directions of research on the one hand (section \ref{subsec:futdir}), and practical applications on the other hand (section \ref{subsec:pracapp}).

\subsection{Future Directions}
\label{subsec:futdir}
One direction for future work consists in improving ASAG further; for this, there are several possible avenues. One aspect of particular interest concerns explainability. In the current setup, ASAG must be considered a black box that transforms inputs (roughly: questions, answers, and reference answers) into outputs (grades). We can evaluate the system’s performance and verify that the outputs do not deviate too much from the desired values in a test set; however, we cannot easily reconstruct how the outputs came to pass, and how much impact the various features of the inputs had on the final prediction.

However, in a high-stakes domain such as grading, explainability is a desirable quality: first, students might demand justifications for the grades they receive. With the current approach, such justifications are hard to generate for individual cases—one would have to make a statistical argument with reference to the training distribution. This makes it hard to respond to student challenges, limiting the applicability of the system. Second, the absence of explainability makes it very difficult to anticipate the system’s performance for inputs not covered in the training and test sets. Such ‘exotic’ inputs might take the form of unintended interactions (for example ‘Please rephrase this question, I do not understand!’), random sequences of symbols, or systematic cheating attempts (for example using adversarial attacks \cite{zouUniversalTransferableAdversarial2023}. These two issues—justification of grades, and robustness with regards to out-of-distribution input—might be addressable by introducing a layer of explainability.

Another way to approach at least the robustness issue is more straightforward: the model’s behaviour for the mentioned input categories can be shaped by simply extending the training and test distributions accordingly, using real or even synthetic data. For example, one might generate random sequences of letters as sample answers, and automatically award a score of zero points. The challenge with this approach is to identify the relevant corner cases – for this, human data exploration (or even red-teaming approaches) seem inevitable. Another addition to this approach would consist in enriching the training dataset with a larger variety of high-complexity questions to enhance the model's grading accuracy and generalizability with the aim to improve its relatively lower performance for questions with higher maximal numbers of points (as mentioned in Section \ref{subsec:expone}, experiment 1).

Another direction for future work consists in extending our human benchmark (Section \ref{subsec:exptwo} above, experiment 2). There are again two avenues which can be pursued: first, we could increase the breadth of the benchmark by increasing the variety of courses and domains. This would allow us to verify that the results presented above generalize, and to study domain-specific differences in performance (for example, we might find that some domains are more ambiguous than others and would hence be able to calibrate our evaluations accordingly). Second, we could increase the depth of our benchmark, obtaining more grades for the dataset at hand. This would allow us to reconstruct the distribution of grades that humans award for the exact same question-answer pair, and compare that distribution to both the grade awarded in the real exam, and the grade awarded by ASAG—how are these situated in the distribution? Are they typical, are they extreme or systematically biased? While both avenues seem promising, the second one—increasing the depth of the benchmark—might unlock more fundamental insights into the grading problem.

\subsection{Practical Applications}
\label{subsec:pracapp}
There are many practical applications for our system—as mentioned above in Section \ref{sec:intro}, grading is one of the most labour-intensive tasks within higher education. Supporting human professionals in this task can improve outcomes for students (by reducing bias and errors), as well as relieve tutors and professors of tedious, repetitive workloads.

However, grades have a high impact—they are used to evaluate learning success, and often form the basis of awarding degrees in higher education. Grading thus strongly affects the lives and careers of individuals and is therefore considered a high-risk domain for the application of technology. This is reflected in recent regulations regarding the use of AI, which places grading-related use cases in the highest risk category \cite{ArtificialIntelligenceAct2023}. 

To unlock the benefits of automated grading while complying with regulations and navigating the associated risks responsibly, we propose to follow the approach taken in the field of autonomous driving. Autonomous driving and automated/autonomous grading share several key features: they are both emerging fields enabled by recent breakthroughs in AI technology, they both operate in high-risk domains, and they both ultimately offer large benefits (both in comfort and safety) via the reduction of human workload and error.

While autonomous driving uses 5 levels (see e.g. \cite{dittySystemsMethodsReliable}), we propose 4 levels for autonomous grading (see Appendix \ref{sec:app:levelauton} for detailed definitions): manual grading (level 0), assisted corrective grading on group level (level 1), assisted corrective grading on single student level (level 2), assisted suggestive grading on single student level (level 3) and autonomous grading (level 4). The grading system described above could in principle be used across all levels. However, at present, legal and academic requirements for the higher levels (in particular level 4) are still evolving, with new regulations emerging as a reaction to trends in technology. We therefore suggest to first focus on level 1 and 2, which focus on assisted corrective grading—a process in which AI-generated grades are merely used to double-check human grades, and flag larger discrepancies for further (human) inspection.

The benefits of this approach are twofold: first, by virtue of being corrective, direct negative impact on students is very unlikely—by design, human graders would not be influenced in their thought process, as the AI grading would happen in parallel behind the scenes. Further, all decisions taken in the process are taken by humans. Second, such processes constitute a formidable data source for further investigations. For example, one might check how often an alert triggered by the AI would in fact lead to an adaptation of the grade, yielding a high-grade evaluation of the AI’s output.

Overall, there are several directions to extend and improve our model, many opportunities to further compare its performance to human graders, and various options for practical applications. However, the results presented above suggest that AI-based grading automation (for example based on models such as ASAG) is a highly promising avenue towards fairer, more consistent and less biased graders for students, while at the same time freeing up the time of tutors for more meaningful teaching interventions, hence benefiting education at large.

\section{Methods}

\subsection{Evaluation Metrics}
\label{subsec:evalmetrics}

\label{sec:methods}
As explained in Section 2, the problem of grading open ended questions can be framed as a regression problem. For evaluating automated grading systems, it is therefore typical to use common regression metrics. Here, we define those in this paper for reference.

\paragraph{1. The Mean Absolute Error (MAE)} reflects differences between paired observations of the same phenomenon (predicted vs true value). It is calculated as the arithmetic average of the absolute errors (deviations of the prediction from the target value). It depends on the units and scaling in the evaluated data set.

\paragraph{2. The Root Mean Square Error (RMSE)} measures the average difference between values predicted by a model and the actual values. It provides an estimation of how well the model can predict the target value (accuracy)—the lower the value, the better the model. The error is squared, and the square root of the mean square deviation is considered. Squaring the error results in strong deviations being weighted more heavily than small ones. The Root Mean Squared Error has the advantage of representing the amount of error in the same unit as the predicted column.

\paragraph{3. The Pearson’s correlation coefficient (Correlation)} 
is the most common way of measuring a linear correlation. It is a number between $-1$ and $1$ that measures the strength and direction of the relationship between two variables. If the value is between $0$ and $1$, there is a positive correlation, meaning if one variable changes, the other variable changes in the same direction. If it is $0$, there is no correlation, e.g. there is no relationship between the variables. If it is between $-1$ and $0$, there is a negative correlation, meaning when one variable changes, the other variable changes in the opposite direction. The Pearson correlation coefficient is the normalized covariance. Therefore, it is independent of the scaling used in the data.

Note that we provide these metrics both measured in absolute points (raw scores ranging between $0$ and $18$ points) and normalized scores (raw score divided by the maximum number of points achievable for the respective exam question, resulting in a percentage). 

\subsection{Data Set Partitioning}
\label{subsec:datapart}

In Section \ref{subsec:setup}, we describe our fine-tuning approach. There, we refer to a partitioning of our dataset into $S_{\text{train}}$, $S_{\text{develop}}$, and $S_{\text{test}}$, with $S_\text{test}$ further divided into $S_{\text{test, unseen questions}}$ and $S_{\text{test, unseen courses}}$. Here, we describe these splits in more detail.

The $S_{\text{train}}$ split, consisting of about 37\% of our proprietary data, forms the backbone of our model's training regimen. It features a diverse range of academic subjects and question complexities encountered in real-world educational settings.

The $S_{\text{develop}}$ split (about 2\% of our data) was derived and subsequently filtered from the $S_{\text{train}}$ data. It was created by taking a random sample of unique questions on a module level (e.g. a set of related courses) and used for model validation and fine-tuning, offering a condensed yet representative cross-section of the training data. This split provides a focused framework for assessing the model's performance on a variety of courses and questions, ensuring robustness and accuracy before proceeding to more challenging unseen data.

The $S_{\text{test}}$ splits (together about 61\% of our data) were used to evaluate model performance as thoroughly as possible, specifically with regards to its ability to generalize to new content. This generalization ability is crucial for real-world applications across varied educational contexts.

The $S_{\text{test, unseen questions}}$ split, encompassing about 60\% of our data, probes the model's ability to handle novel questions from both seen and unseen courses. It presents an expansive array of questions to test the model's versatility and adaptability across a broad spectrum of academic disciplines and unseen questions.

Finally, the $S_{\text{test, unseen courses}}$ (together about 1\% of our data) sub-split tests the model capabilities to grade entirely new courses not represented in the training or development data. This held-out split tests the model generalization and transfer-learning capabilities in grading unfamiliar content and handling new grading criteria encountered in new courses.

\section{Acknowledgement}
\label{sec:ack}
The authors would like to thank the AI team at IU International University of Applied Sciences, in particular Valerie Hekkel, Alena Vasilevich and Amos Schikowsky, for valuable discussions and constructive feedback. We further want to acknowledge Michael Mohler for providing a grading dataset gathered at the University of North Texas.

%\bibliographystyle{./unsrt2authabbrvpp}  
%\bibliographystyle{plain}
%\bibliography{ASAG_Bibliography} 
\printbibliography[title={References}]

%%%%%%%%% APPENDIX %%%%%%%%%%%%%%%%%%%%%%%%%%%%%%%%%%%%%%%%%%%%%%%%%%%%%%%%%%%%%%%%%%%%%%%%%%%%%%%%%%%%%%
\pagebreak
\textbf{\large Appendix}
%%%%%%%%%% Prefix a "A" to all figures, tables and reset the counter %%%%%%%%%%
\setcounter{section}{0}
\setcounter{figure}{0}
\setcounter{table}{0}
\makeatletter
\renewcommand{\thesection}{A\arabic{section}}
\renewcommand{\thefigure}{A\arabic{figure}}
\renewcommand{\thetable}{A\arabic{table}}

\section{Evaluating our model on the dataset of Mohler 2011 - c.f. \cite{mohlerLearningGradeShort2011}}
\label{sec:app:evalmohler}

\begin{table}[H]
    \centering
    \caption{Comparison of Models}
    \label{tab:app:mohlerperf}
    \begin{tabular}{lccc}
        \toprule
        & \textbf{Mohler 2011} & \textbf{Schlippe 2022} & \textbf{Our Model} \\
        \midrule
        \textbf{RMSE (points)} & 0.998 & 0.72 & 1.511 \\
        \textbf{Pearson (points)} & 0.518 & 0.73 & 0.602 \\
        \textbf{Mean Absolute Error (points)} & - & 0.42 & 0.890 \\
        \bottomrule
    \end{tabular}
\end{table}

\section{Courses used for re-grading}
\label{sec:app:regrading}

\begin{table}[H]
    \centering
    \caption{Regrading evaluation on course level. RMSE and Pearson’s correlation coefficient (normalized values)}
    \label{tab:app:coursecomp}
    \begin{tabular}{lcccc}
        \toprule
        \textbf{Course} & \makecell{\textbf{RMSE} \\ human \\ vs. \\ human} & \makecell{\textbf{RMSE} \\ human \\ vs. \\ model} & \makecell{\textbf{Pearson} \\ human \\ vs. \\ human} & \makecell{\textbf{Pearson} \\ human\\ vs. \\\ model} \\
        \midrule
        Marketing II & 0.303 & 0.297 & 0.701 & 0.561 \\
        Personal und Unternehmensführung & 0.328 & 0.277 & 0.607 & 0.660 \\
        Diversity Management & 0.371 & 0.330 & 0.228 & 0.237 \\
        Differentielle und Persönlichkeitspsychologie & 0.245 & 0.231 & 0.747 & 0.741 \\
        Sozialpsychologie & 0.360 & 0.243 & 0.464 & 0.664 \\
        Besonderes Schuldrecht I & 0.315 & 0.282 & 0.579 & 0.650 \\
        Grundlagen des bürgerlichen Rechts I & 0.381 & 0.330 & 0.412 & 0.321 \\
        Öffentliches Recht I & 0.214 & 0.213 & 0.687 & 0.718 \\
        Schuldrecht I, Einführung & 0.294 & 0.326 & 0.619 & 0.538 \\
        Leadership & 0.358 & 0.223 & 0.714 & 0.796 \\
        Globale Unternehmen und Globalisierung & 0.532 & 0.276 & 0.461 & 0.5688 \\
        Artificial Intelligence & 0.5401 & 0.259 & 0.397 & 0.650 \\
        Data Utilization & 0.506 & 0.358 & 0.466 & 0.611 \\
        Data Science & 0.583 & 0.282 & 0.308 & 0.580 \\
        Change Management und Organisationsentwicklung & 0.449 & 0.301 & 0.493 & 0.275 \\
        Machine Learning & 0.557 & 0.278 & 0.413 & 0.623 \\
        \bottomrule
    \end{tabular}
\end{table}

\begin{figure} 
    \centering
    \includegraphics[width=.8\textwidth]{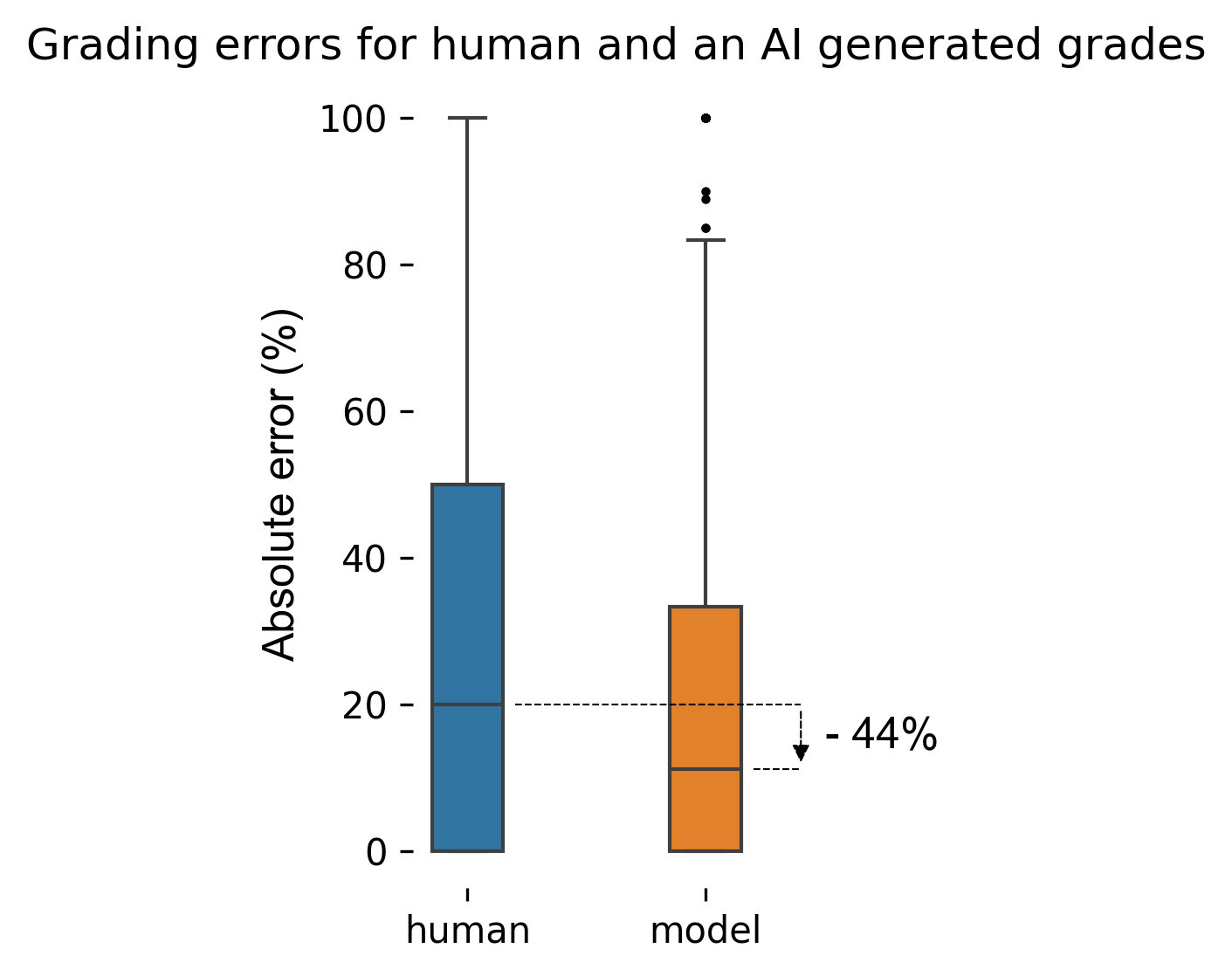}
    \caption[Error distributions from the human benchmark experiment] {\textbf{Error distributions from the human benchmark experiment.} We show the distribution of absolute grading errors for both the human regraders (left, blue) and the model (right, orange). The median absolute errors are 20.0 and 11.1 respectively. The model’s median absolute error is hence 44\% smaller than the humans’ median absolute error. The distributions differ significantly (p > 0.001, Mann-Whitney U test).}
    \label{fig:app:graderror}
\end{figure}

\section{Detailed results of deviation comparison}
\label{sec:app:detres}
\begin{table}[H]
    \caption{Comparison of official human grader versus regrader, official human grader versus model and average of both human graders versus model, all regraders included (1600 rows).}
    \label{tab:app:gradvsregradone}
    \centering
    \begin{tabular}{lccc}
        \toprule
        \textbf{Metric} & \textbf{Human vs. Human} & \textbf{Human vs. Model} & \textbf{Average of Humans vs. Model} \\
        \midrule
        \textbf{MAE (points)} & 2.862 & 1.789 & 2.599 \\
        \textbf{RMSE (points)} & 4.566 & 3.061 & 4.171 \\
        \textbf{Correlation (points)} & 0.583 & 0.761 & 0.571 \\
        \midrule
        \textbf{MAE (normalized)} & 0.289 & 0.183 & 0.265 \\
        \textbf{RMSE (normalized)} & 0.412 & 0.284 & 0.378 \\
        \textbf{Correlation (normalized)} & 0.485 & 0.590 & 0.511 \\
        \bottomrule
    \end{tabular}
\end{table}

\begin{table}[H]
    \caption{Comparison of official human grader versus regrader, official human grader versus model and average of both human graders versus model, extreme regraders excluded (1000 rows).}
    \label{tab:app:gradvsregradtwo}
    \centering
    \begin{tabular}{lccc}
        \toprule
        \textbf{Metric} & \textbf{Human vs. Human} & \textbf{Human vs. Model} & \textbf{Average of Humans vs. Model} \\
        \midrule
        \textbf{MAE (points)} & 1.888 & 1.465 & 1.696 \\
        \textbf{RMSE (points)} & 2.926 & 2.346 & 2.579 \\
        \textbf{Correlation (points)} & 0.643 & 0.741 & 0.671 \\
        \midrule
        \textbf{MAE (normalized)} & 0.221 & 0.178 & 0.204 \\
        \textbf{RMSE (normalized)} & 0.321 & 0.279 & 0.296 \\
        \textbf{Correlation (normalized)} & 0.589 & 0.600 & 0.613 \\
        \bottomrule
    \end{tabular}
\end{table}

\section{Levels of autonomous grading}
\label{sec:app:levelauton}
Here, we provide a more detailed description of the levels of autonomous grading by listing use cases for each level. 
\subsection*{Level 0: Manual grading}
\label{subsec:app:levelzero}
No use cases here; all tasks are performed by humans; no AI support is used.

\subsection*{Level 1: Assisted grading (on student group level)}
\label{subsec:app:levelone}
\paragraph{Use case 1.1 - Detect outlier graders:} A human grades exams, as per the current process. After grading, all exams are graded again by the AI model in the background. The human and AI grades are then compared, the differences are aggregated across all exams. If the difference is too large, an alert is triggered, and another human (for example, a more senior academic) is tasked to review the grades. Note that \textit{individual AI-generated grades are never shown} to anyone involved—the only information that leaves the system is an alert if deviations are too large on average. The benefit of this system is to protect students from bad grading and protect academic staff from complaints.

\subsection*{Level 2: Assisted corrective grading (on single-student level)}
\label{subsec:app:leveltwo}
\paragraph{Use case 2.1 - Detect outlier question scores:} Like the above ‘detect outlier graders’, but with higher resolution—now, alerts contain reports that \textit{highlight single questions with high deviations between human and AI} and escalate those to another human (for example, a more senior academic). This saves humans additional work, as they can review problematic questions directly, and it can also improve outcomes, as it prevents human errors in the review step.

\subsection*{Level 3: Assisted suggestive grading}
\label{subsec:app:levelthree}
\paragraph{Use case 3.1 - Pre-fill feedback:} A human grades exams, as per the current process. After the human enters a score (for example, 5/10 points), an AI model generates an explanation (for example, “You correctly mentioned some reasons …. However, you did not mention …. Therefore, you receive half of the maximum score.”). The \textit{generated explanation is then suggested to the human grader}, who can decide to use it, modify, or discard it. This substantially speeds up the grading process, and potentially improves results, as AI-generated feedback might be more detailed.

\subsection*{Level 4: Autonomous grading}
\label{subsec:app:}
\paragraph{Use case 4.1 - Degrees without exams:}  In some courses, IU currently offers a function called ‘Trainer’, which helps students to practise what they learned (see Fig A1 for examples). Students use this app during their studies, and steadily progress. \textit{The AI model grades their answers, and slowly builds up the capability to predict whether a student would answer a question correctly or not.} This capability is then used to predict a student’s performance in a \textit{hypothetical exam} -- it predicts the grade a student would have if they took the exam today. After sufficient usage, and upon reaching the desired grade across all course contents, the predicted overall grade is used to certify the student’s knowledge and award the degree. Upon request, humans can review the grades before final submission, or administer a traditional exam for those that insist on it. This system reduces human grading effort to nearly zero, saving teachers tremendous amounts of time. Further, it reduces human errors and biases from grading, making grades more just and accurate. Finally, students are spared the anxiety and stress that often accompany exams—instead, they are evaluated much more broadly across time, yielding a much more comprehensive and reliable outcome.

\begin{figure}[H] 
    \centering
    \includegraphics[width=.5\textwidth]{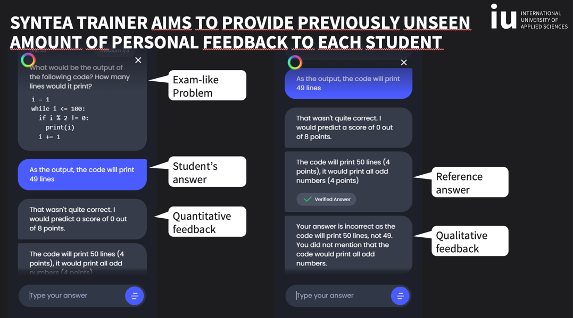}
    \caption[The exam trainer feature] {\textbf{The exam trainer feature.} At IU International University of Applied Sciences, students can practise for their exams using an AI-powered app.}
    \label{fig:app:syntea}
\end{figure}

\end{document}